\documentclass[conference]{IEEEtran}
\IEEEoverridecommandlockouts
\usepackage{cite}
\usepackage{hyperref}
\usepackage{amsmath,amssymb,amsfonts}
\usepackage{graphicx}
\usepackage{textcomp}
\usepackage{algorithm}
\usepackage{algpseudocode}
\usepackage{subcaption}
\usepackage{inputenc}
\usepackage{xcolor}
\usepackage{soul}
\def\BibTeX{{\rm B\kern-.05em{\sc i\kern-.025em b}\kern-.08em
    T\kern-.1667em\lower.7ex\hbox{E}\kern-.125emX}}

\newcommand{\commentbyKG}[1]{}

\begin{document}

\title{Joint Probability Estimation Using Tensor Decomposition and Dictionaries
\thanks{AR wishes to thank SERB Matrics Grant MTR/2019/000691.}
}

\author{\IEEEauthorblockN{Shaan ul Haque}
\IEEEauthorblockA{\textit{Department of EE} \\
\textit{IIT Bombay}\\
Mumbai, India \\
shaanhaque2016@gmail.com}
\and
\IEEEauthorblockN{Ajit Rajwade}
\IEEEauthorblockA{\textit{Department of CSE} \\
\textit{IIT Bombay}\\
Mumbai, India \\
ajitvr@cse.iitb.ac.in}
\and
\IEEEauthorblockN{Karthik S. Gurumoorthy}
\IEEEauthorblockA{\textit{Amazon} \\
Bengaluru, India \\
karthik.gurumoorthy@gmail.com}
}

\maketitle

\begin{abstract}
In this work, we study non-parametric estimation of joint probabilities of a given set of discrete and continuous random variables from their (empirically estimated) 2D marginals, under the assumption that the joint probability could be decomposed and approximated by a mixture of product densities/mass functions. The problem of estimating the joint probability density function (PDF) using semi-parametric techniques such as Gaussian Mixture Models (GMMs) is widely studied. However such techniques yield poor results when the underlying densities are mixtures of various other families of distributions such as Laplacian or generalized Gaussian, uniform, Cauchy, etc. Further, GMMs are not the best choice to estimate joint distributions which are hybrid in nature, i.e., some random variables are discrete while others are continuous. We present a novel approach for estimating the PDF using ideas from dictionary representations in signal processing coupled with low rank tensor decompositions. To the best our knowledge, this is the first work on estimating joint PDFs employing dictionaries alongside tensor decompositions. We create a dictionary of various families of distributions by inspecting the data, and use it to approximate each decomposed factor of the product in the mixture. Our approach can naturally handle hybrid $N$-dimensional distributions. We test our approach on a variety of synthetic and real datasets to demonstrate its effectiveness in terms of better classification rates and lower error rates, when compared to state of the art estimators. 
\end{abstract}

\begin{IEEEkeywords}
Statistical Learning, Tensor Decomposition, Dictionary Representation, Probability Density Estimation
\end{IEEEkeywords}

\section{Introduction}
Inferring probability density functions (PDFs) from data is a fundamental problem in machine learning, statistics and signal processing\cite{Murphy2012}\cite{Wainwright2019} with applications in varied fields such as conditional inference, samples generation, image reconstruction and many more. The applicability of widely popular methods like Gaussian mixture models (GMMs) are restricted to the case of smooth multi-modal densities where every mode is well approximated by Gaussians. Similar is the case with non-parametric settings such as kernel density estimation (KDE). Moreover, such techniques exhibit lower and lower convergence rates as the data dimensionality increases. For $N$-dimensional data, the convergence rate, in terms of the integrated mean square error (MISE), for the KDE is known to be $\mathcal{O}(N^{-\frac{4}{N+4}}_s)$\cite{Chen2018}.\\
Recently, joint probability mass functions (PMFs) of discrete or discretized random variables (RVs) have been represented as tensors -- in fact \emph{low rank} tensors, using the fact that the different RVs are neither completely dependent nor completely independent \cite{Kargas2018}. There have been significant developments in the estimation of joint PMFs from lower dimesional marginals (eg, 3D marginals) using the Canonical Polyadic Tensor Decompostion (CPD) using these low rank constraints \cite{Kargas2018}. Along similar lines, the work in \cite{Ibrahim2021} uses non-negative matrix factorization (NMF) techniques to estimate the PMF from just pairwise (i.e. 2D) marginals. On the other hand, ideas from tomography were incorporated into this tensor-based framework in \cite{Vora2021} for PMF reconstruction from just 1D marginals. Extending these ideas to the continuous domain, one can discretize the continuous RVs, use the aforementioned techniques for estimating cumulative interval measures (CIMs), followed by an appropriate interpolation technique to recover the joint PDF. For instance, \cite{Kargas2019} uses sinc interpolation assuming that the underlying PDF is band-limited, in keeping with the popular Shannon-Nyquist theorem. Under similar assumptions, there also exists work in the Fourier domain where CIM reconstruction techniques are applied to obtain the `characteristic tensor' and the continuous PDF is then retrieved using the inverse Fourier Transform \cite{Amiridi2020}.\\
In this work, we present a novel approach which combines ideas from the CPD model for tensors and dictionary representations, to reconstruct the joint PDF from just pairwise (2D) marginals, as opposed to 3D marginals. The key idea is to prepare a dictionary of 1D PDFs belonging to various families, with parameters restricted to lie in a carefully chosen range, for each component of the $N$-dimensional data. Reconstruction of the $N$-dimensional PDF using such a dictionary helps us to circumvent restrictive assumptions such as PDF smoothness or band-limitedness as in previous methods. Furthermore, the convergence rate for estimation of 2D marginals is superior to that for 3D marginals used in \cite{Kargas2018}. 

\section{Background}
\subsection{Canonical Polyadic Decomposition(CPD) of Tensors}
Any $N$-dimesional tensor $\boldsymbol{\underline{T}}\in \mathbb{R}^{I_1\times I_2\times...\times I_N}$ admits a decomposition in the form of the sum of $F$ rank-1 tensors. This is known as the CPD, and is given by:
\begin{equation}\label{CPD}
   \boldsymbol{\underline{T}}=\sum_{r=1}^F \boldsymbol{\lambda}[r] \boldsymbol{A_1}[:,r]\circ \boldsymbol{A_2}[:,r]\circ....\boldsymbol{A_N}[:,r],
\end{equation}
where $F$ is the smallest number for which such decomposition is possible, where for each $n \in [N]$, where $[N] \triangleq [1,2,...,N]$ the matrix $\boldsymbol{A_n} \in \mathbb{R}^{I_n\times F}$ is called a mode factor matrix, $\circ$ denotes the outer product of vectors and $\lambda[r]$ denotes the $r^{\textrm{th}}$ mixing weight. For tensors that represent high-dimensional PMFs, the above decomposition is applicable with the following additional constraint: (1) $\forall n \in [N], r \in [F], \|\boldsymbol{A_n}[;,r]\|_1=1$ with non-negative entries in the mode factor matrices, and (2) $\|\boldsymbol{\lambda}\|_1 = 1$ where $\boldsymbol{\lambda}$ is a vector of $F$ non-negative mixing weights. Recovering the PMF is equivalent to estimating these mode factors and $\boldsymbol\lambda$\cite{Kargas2018}. \\
The above model can be viewed as a naive Bayes model with the latent variable $H$ that takes on $F$ different values such $P(H=r)=\lambda[r]$ for all $r \in [F]$, and the outer product of the mode factor can be viewed as a conditional probability given $H$ \cite{Kargas2018}. With this, the CPD model for the PDF of RV $\boldsymbol{X} = (X_1, X_2, ..., X_N) \in \mathbb{R}^N$ can be formulated as:
\begin{equation}\label{Bayesian_Model}
   \boldsymbol{\underline{T}}=\sum_{r=1}^F P(H=r)\prod_{n=1}^{n=N}P(X_n=i_n|H=r).
\end{equation}
\subsection{Continuous RVs: PDF estimation}
For the case of continuous RVs, consider an $N$-dimensional RV $\boldsymbol{X}=\{X_n\}_{n=1}^N$, whose PDF is given by the following mixture of multivariate distributions:
\begin{equation}\label{Cont_case}
    f_{\boldsymbol{X}}(x_1,...x_N)=\sum_{r=1}^F \boldsymbol{\lambda}[r]f_{\boldsymbol{X}|H}(x_1,...,x_N|H=r).
\end{equation}
If the RVs are independent given $H$, then each conditional density can be represented by the product of 1D densities and the above equation becomes:
\begin{equation}\label{Cont_Bayesian_Model}
    f_{\boldsymbol{X}}(x_1,...,x_N)=\sum_{r=1}^F \boldsymbol{\lambda}[r]\prod_{n=1}^N f_{X_n|H}(x_n|H=r),
\end{equation}
which can be viewed as the continuous analog of the CPD model for PMFs\cite{Kargas2019}. In this case, the problem of estimating the PDF is equivalent to estimating these `continuous' mode factors and their mixing weights.

\subsection{Joint PDF estimation from 3D marginals}
The work in \cite{Kargas2019} exploits the above CPD of densities which are conditionally independent to estimate the joint PDF from the data. They propose to discretize each component $X_n$ of the $N$-dimensional variable into $I_n$ intervals $\{\Delta_n^i \triangleq (d_n^{i-1}, d_n^i)\}_{1\leq i\leq I_n}$ and form the CIM tensor $\boldsymbol{\underline{Z}}$ given by:
\begin{align}\label{PMF}
    \boldsymbol{\underline{Z}}(i_1,...,i_N)&=P(X_1\in \Delta_1^{i_1},..., X_N\in \Delta_N^{i_N})\\
    &=\sum_{r=1}^F \boldsymbol{\lambda}[r]\prod_{n=1}^N P(X_n\in \Delta_n^i|H=r),
\end{align}
where $P(X_n\in \Delta_n^i|H=r)$ are the CIMs of the corresponding 1D components of the RV. If the PDF/CDF of each component in the product is band-limited, the CIMs can be estimated using the PMF reconstruction techniques described in \cite{Kargas2018}. The reconstruction involves estimating the 3D marginals, $\boldsymbol{\underline{Z}}_{i,j,k}=P(X_i,X_j,X_k)$, using standard histogramming and then minimizing the following cost function:
\begin{gather}
    \min_{\boldsymbol{\{A_n\}}_{n=1}^N, \boldsymbol\lambda}\sum_i\sum_{j>i}\sum_{k>j}\|\boldsymbol{\underline{Z}}_{i,j,k}-[\boldsymbol{\lambda}, \boldsymbol{A_i},\boldsymbol{A_j},\boldsymbol{A_k}]\|_F^2 \nonumber\\
    \label{eq:Zijkcostfunction}
    \textrm{ s.t. } \ \forall n,r \ \|\boldsymbol{A_n}[:,r]\|_1=\|\boldsymbol{\lambda}\|_1=1, \boldsymbol{A_n} \succeq \boldsymbol{0}, \boldsymbol{\lambda} \succeq \boldsymbol{0},
\end{gather}
where $\succeq$ represents the element-wise inequality and $[\boldsymbol{\lambda}, \boldsymbol{A_i},\boldsymbol{A_j},\boldsymbol{A_k}] \triangleq \sum_{r=1}^F \lambda[r]\boldsymbol{A_i}[:,r]\circ \boldsymbol{A_j}[:,r]\circ\boldsymbol{A_k}[:,r]$. After obtaining the discretized samples of the PDFs, the technique in \cite{Kargas2019} invokes the Shannon-Nyquist sampling theorem to use sinc interpolation to produce the original joint PDF using Eqn.~\ref{CPD}. \\
In \cite{Amiridi2020}, under similar band-limitedness of the density, a different kind of approach is considered using the characteristic function of the density, $\boldsymbol{\Phi_{X}}(\boldsymbol\nu)=E[e^{j\boldsymbol{\nu}^T \boldsymbol{X}}],$
where $j \triangleq \sqrt{-1}$. If the true density is given by $f_{\boldsymbol X}(\boldsymbol x)$, then it can be well approximated by truncating the Fourier series below:
\begin{gather}\label{FT}
    \hat{f}_{\boldsymbol{X}}(\boldsymbol {x})=\sum_{k_1=-K_1}^{k_1=K_1}...\sum_{k_N=-K_N}^{k_N=K_N}\boldsymbol{\Phi_{X}}(\boldsymbol{k})e^{-2\pi j\boldsymbol{k}^T \boldsymbol {x}}.
\end{gather}
If we furthur impose the CPD model on it, the expression for $\boldsymbol{\underline{\Phi}_{X}}(\boldsymbol{k})$ becomes becomes similar to Eq. \ref{Bayesian_Model}.
where $F$ is the rank of the tensor. With enough samples $\{\boldsymbol{x}_m\}$, the expectation can be reliably estimated using sample mean $ \boldsymbol{\hat{\underline{\Phi}}_{X}}(\boldsymbol\nu)=\frac{1}{M}\sum_{m=1}^Me^{j\boldsymbol{\nu}^T \boldsymbol{x_m}}$
Minimizing a cost function similar to Eqn.\ref{eq:Zijkcostfunction} for the characteristic tensor $\boldsymbol{\hat{\underline{\Phi}}_{X}}(\boldsymbol\nu)$ followed by inverse Fourier transform, yields $f_{\boldsymbol X}(\boldsymbol x)$.

\subsection{Joint PMF estimation from 2D marginals}
An interesting approach that employs Non-negative Matrix Factorization (NMF) techniques\cite{NMF} to estimate the mode factors from 2D marginals was introduced in \cite{Ibrahim2021}. They estimate $\boldsymbol{Z}_{j,k}$ via sample histogramming and obtain the mode factors using the relation $\boldsymbol{Z}_{j,k}=\boldsymbol{A_j \Lambda A_k}^T$, where $\boldsymbol{\Lambda}$ is a diagonal matrix with diagonal elements obtained from $\boldsymbol{\lambda}$. If the tensor rank $F \gg \textrm{min}(I_j,I_k)$, then NMF techniques cannot be applied \cite{Fu2018}. Therefore, the authors proposed to split the indices of $N$ variables into two sets and construct a matrix $\boldsymbol{\tilde{Z}}$ by row and column concatenation using the indices in the two sets (see \cite[Eqn. 3]{Ibrahim2021}. Then $\boldsymbol{\tilde{Z}}$ is decomposed as $\boldsymbol{\tilde{Z}}=\boldsymbol{WH}^T$ using the successive projection algorithm (SPA) \cite{Gillis2014}. The mode factors are then extracted using the relations $\boldsymbol{W}=[\boldsymbol{A}_{l_1}, \boldsymbol{A}_{l_2},..., \boldsymbol{A}_{l_M}]^T$ and $\boldsymbol{H}^T=\boldsymbol\Lambda[\boldsymbol{A}_{l_{M+1}}, \boldsymbol{A}_{l_{M+2}},..., \boldsymbol{A}_{l_N}]$, where $\{l_1, l_2,...,l_M\}$ and $\{l_{M+1}, l_{M+2}..., l_N\}$ are the two sets of indices.

\section{Problem Statement and Algorithm}
Let $\boldsymbol{X} \triangleq (X_1, X_2,...,X_N)$ be an $N$-d RV, each component of which can be either continuous or discrete. Our aim is to estimate $f_{\boldsymbol{X}}(\boldsymbol{x})$ which follows the CPD model with rank $F$ for the continuous case, given sample values of the RV. Let us further assume that each column of these ``continuous mode factors" $f_{X_n|H}(x_n|H=r)$ are convex combinations of various densities from a given dictionary. Mathematically, we have $f_{X_n|H}(x_n|H=r)=\boldsymbol{\mathcal{A}_n}[:,r] = \boldsymbol{\mathcal{D}_n}\boldsymbol{B_n}[:,r]$, $1\leq r\leq F$, where $\boldsymbol{\mathcal{D}_n}$ is a dictionary of continuous densities (or discrete PMFs in some cases) and $\boldsymbol{B_n}[:,r]\in \mathbb{R}_+^{L_n}\cup\{\boldsymbol{0}\}$ is the non-negative weight vector which sums to 1. 
Here $L_n$ is the number of different densities (number of columns) that are present in the dictionary $\boldsymbol{\mathcal{D}_n}$. 
We will later see that keeping the dictionary $\boldsymbol{\mathcal{D}_n}$ separate for each component of the RV gives us a lot of flexibility in dealing with RVs defined on disparate domains. Then, the 2D PDFs $\boldsymbol{\mathcal{Z}_{j,k}}$ can be expressed in the form $\boldsymbol{\mathcal{Z}_{j,k}}=\boldsymbol{\mathcal{D}_j}\boldsymbol{B_j}\boldsymbol{\Lambda}\boldsymbol{B_k}^T\boldsymbol{\mathcal{D}_k}^T$ which is straightforward to derive by marginalizing along $X_j$ and $X_k$, and replacing each density in the product by its dictionary representation. We discretize each component of the RV into $I_n$ intervals $\{\Delta_n^i \triangleq (d_n^{i-1}, d_n^i)\}_{1\leq i\leq I_n}$ and form the 2D PMF matrix $\boldsymbol{\bar{Z}_{j,k}}$ given by:
\begin{align}\label{2D_PMF}
    \boldsymbol{\bar{Z}_{j,k}}(i_j, i_k)&=P(X_j\in \Delta_j^{i_j}, X_k\in \Delta_k^{i_k})\\ \nonumber
    &=\boldsymbol{\bar{D}_j}[i_j,:]\boldsymbol{B_j}\Lambda\boldsymbol{B_k}^T\bar{\boldsymbol{D}}_k[i_k,:]^T,
\end{align} 
where the column-wise discretized form of a ``continuous tensor/matrix" $\boldsymbol{\mathcal{Z}}$ is represented as $\boldsymbol{\bar{Z}}$. We can obtain an estimate $\boldsymbol{\hat{Z}_{j,k}}$ of these 2D PMFs by standard histogramming of the samples. It is emphasised here that we do not assume any knowledge of the densities from which the data are generated. For the purpose of estimation, all the dictionaries $\{\boldsymbol{\mathcal{D}}_n\}_{n=1}^N$ are designed by inspecting the samples. We will elaborate on this aspect in more detail in Sec.\ref{sec:exp}. Thus, our task is to estimate the coefficients $\boldsymbol{\Lambda}, \boldsymbol{B}_1,...,\boldsymbol{B_n}$. To this end, we minimize the following cost function:
\begin{gather}\label{eq:jupad_cost_fn}
    J(\{\boldsymbol{B_n}\}_{n=1}^N, \boldsymbol{\Lambda})=\sum_{j, j<k}\|\boldsymbol{\hat{Z}_{j,k}}-\boldsymbol{\bar{D}_j}\boldsymbol{B_j}\boldsymbol{\Lambda}\boldsymbol{B_k}^T\boldsymbol{\bar{D}_k}^T\|_F^2\\ \nonumber
     \textrm { s.t. } \forall j,r, \|\boldsymbol{B_j}[:,r]\|_1 = 1, \boldsymbol{B_j} \succeq \boldsymbol{0},
     \|\textrm{diag}(\boldsymbol{\Lambda})\|_1 = 1, \boldsymbol{\Lambda} \succeq \boldsymbol{0}.
\end{gather}
However, minimization via (say) a simple gradient descent may not be feasible as the solution will not be identifiable for cases when $F>\textrm{min}(L_j,L_k)$, as argued in \cite{Ibrahim2021}. We propose to do the following: (i) We minimize $\|\hat{\boldsymbol{Z}}_{j,k}-\boldsymbol{\bar{D}_j}\boldsymbol{T}_{j,k}\boldsymbol{\bar{D}_k}^T\|_F^2$ for each pair of $(j,k)$ via mirror descent to obtain $\boldsymbol{T}_{j,k} \triangleq \boldsymbol{B}_j \boldsymbol{\Lambda} \boldsymbol{B}_k^T$. For mirror descent, closed-form updates can be derived even with the simplex constraint on $\boldsymbol{T}_{j,k}$ \cite{Kivinen1997}\cite{Beck2003}. (ii) Next, we construct the matrix $\Tilde{\boldsymbol{T}}$ from $\boldsymbol{T}_{j,k}$ using the concatenation approach described in \cite[Eqn. 3]{Ibrahim2021} and determine the matrices $\boldsymbol{W}$ and $\boldsymbol{H}^T$ as outputs from function SPA(.) implemented in \cite{Ibrahim2021}. The weight matrices, $\boldsymbol{B_n}$ and $\boldsymbol{\Lambda}$ can now be identified as submatrices of $\boldsymbol{W}$ and $\boldsymbol{H}^T$. (iii) We further refine our estimates for these weights by using mirror descent on cost function in Eqn.~\ref{eq:jupad_cost_fn}. We name our algorithm \textbf{JUPAD}: \textbf{J}oint density estimation \textbf{U}sing \textbf{P}airwise marginals \textbf{A}nd \textbf{D}ictionaries.\\
The pseudo-code for the algorithm is presented in Alg.\ref{alg:cap}, where $\eta_T, \eta_B$ and $\eta_L$ are the learning rate hyper-parameters chosen via cross-validation, $\otimes$ represents the Hadamard product of two matrices, and $\textrm{vec}(.)$ reshapes a matrix into vector form.

\begin{algorithm}
\caption{Joint probability density estimation using pairwise marginals}\label{alg:cap}
\begin{algorithmic}[1]
\State \textbf{Procedure: JUPAD}
\State Obtain the estimate $\hat{\boldsymbol{Z}}_{j,k}$ for the 2D marginals $\boldsymbol{Z}_{j,k}$ via histogramming
\For{each pair $(j,k), \ j<k$}
\State Randomly initialize $\boldsymbol{T}_{j,k}$ 
\While {converged==false}
\State $\boldsymbol{T}_{j,k} \gets \boldsymbol{T}_{j,k}\otimes exp(-\eta_T\frac{\partial(\|\hat{\boldsymbol{Z}}_{j,k}-\bar{\boldsymbol{D}}_j\boldsymbol{T}_{j,k}\bar{\boldsymbol{D}}_k^T\|_F^2)}{\partial \boldsymbol{T}_{j,k}})$
\State $\boldsymbol{T}_{j,k} \gets \frac{\boldsymbol{T}_{j,k}}{\|\textrm{vec}(\boldsymbol{T}_{j,k})\|_1}$
\EndWhile
\EndFor
\State Assemble $\Tilde{\boldsymbol{T}}$ from $\{\boldsymbol{T}_{j,k}\}$ following \cite[Eqn. 3]{Kargas2019}.
\State $\{\boldsymbol{B_n}\}_{n=1}^N$, $\boldsymbol\Lambda \gets SPA(\Tilde{\boldsymbol{T}})$
\While{converged==false}
\For {i=1 to N}
\While{converged==false}
\State $\boldsymbol{B}_n \gets \boldsymbol{B}_n\otimes exp(-\eta_B\frac{\partial J}{\partial \boldsymbol{B}_n})$
\State $L_1$ normalize each column of $\boldsymbol{B}_n$
\EndWhile
\EndFor
\While{converged==false}
\State $\boldsymbol\Lambda \gets \boldsymbol\Lambda\otimes exp(-\eta_L\frac{\partial J}{\partial \boldsymbol\Lambda})$
\State $L_1$ normalize the diagonal of $\boldsymbol\Lambda$
\EndWhile
\EndWhile
\end{algorithmic}
\end{algorithm}

\section{Numerical Results}
\begin{figure*}[t]
    \begin{subfigure}[t]{.5\linewidth}
	    \centering
        \includegraphics[width=60mm,height=35mm]{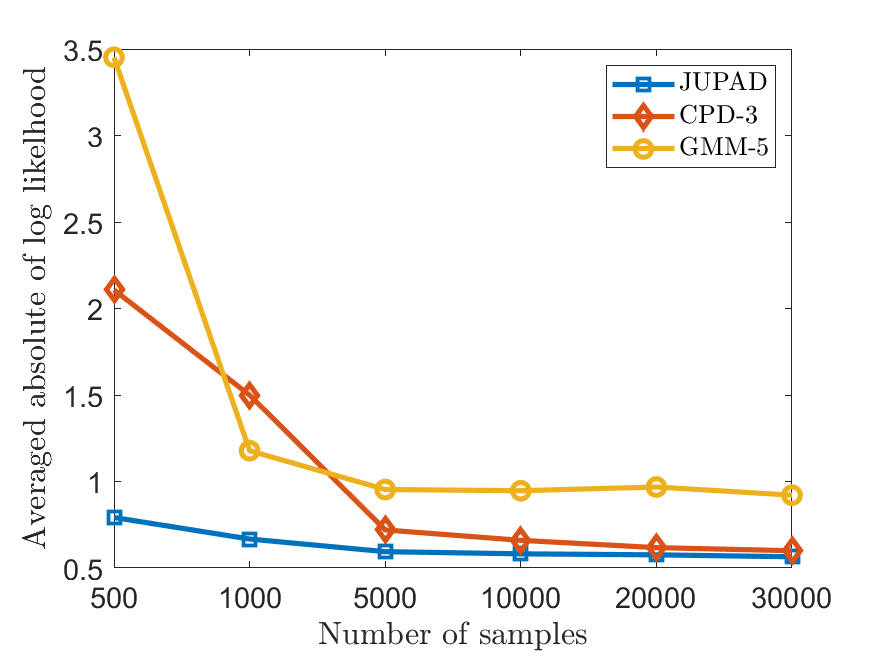}
        \caption{Mixture of Laplacians}
        \label{fig:exp1}
    \end{subfigure}	
    \begin{subfigure}[t]{.5\linewidth}
	    \centering
        \includegraphics[width=60mm,height=35mm]{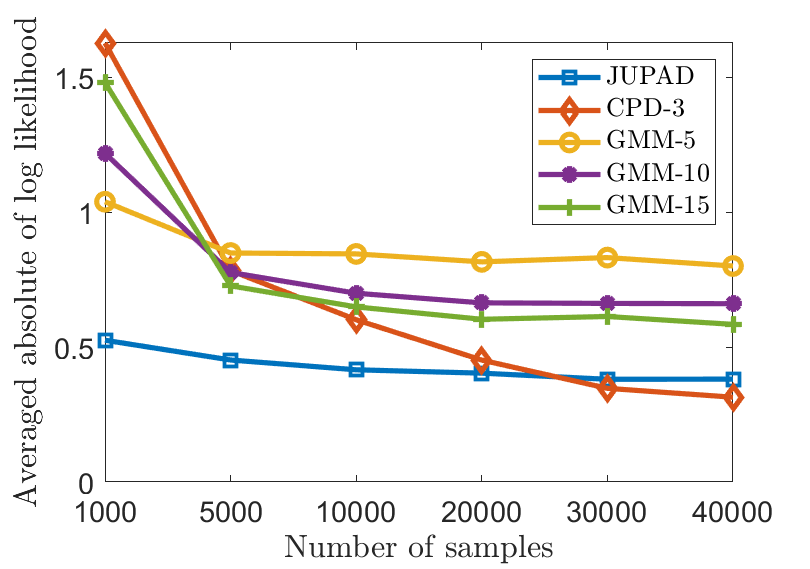}
        \caption{Mixture of Gaussians}
        \label{fig:exp2}
    \end{subfigure}	
	\begin{subfigure}[t]{.5\linewidth}
	    \centering
        \includegraphics[width=60mm,height=35mm]{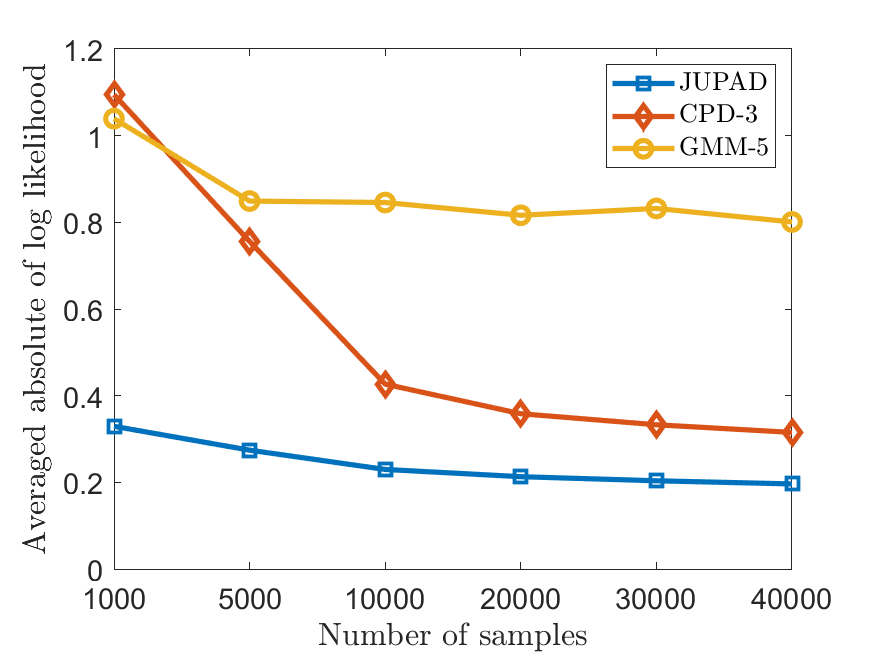}
        \caption{Mixture of Gaussians and Laplacians}
        \label{fig:exp3}
    \end{subfigure}	
    \begin{subfigure}[t]{.5\linewidth}
	    \centering
        \includegraphics[width=60mm,height=35mm]{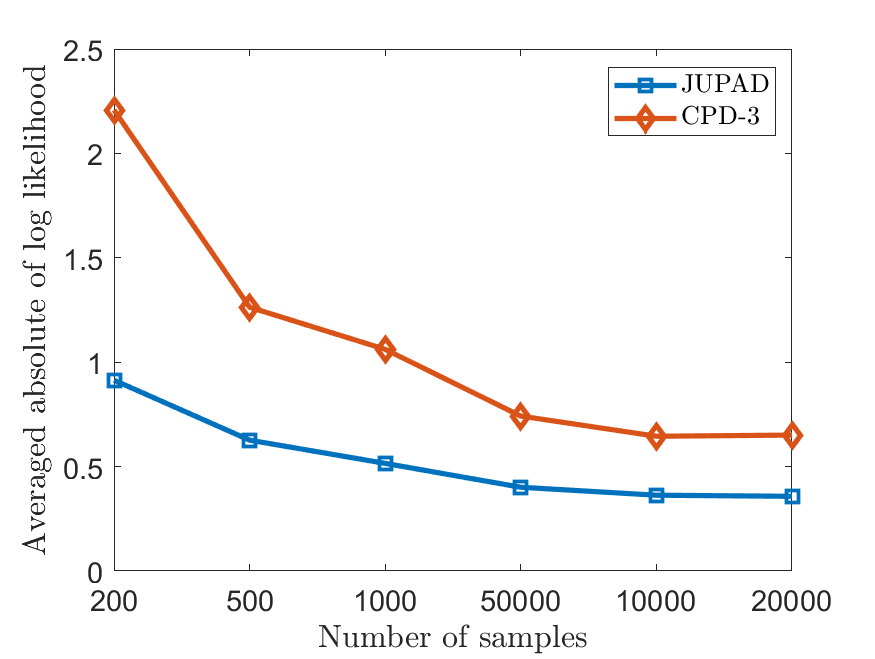}
        \caption{RV with continuous and discrete components}
        \label{fig:exp4}
    \end{subfigure}	
\caption{$D(\widehat{f}_{\boldsymbol{X}}, f_{\boldsymbol{X}})$ vs Number of samples ($N_s$) for JUPAD (our approach), CPD-3 \cite{Kargas2019} and GMM.}
\end{figure*}
\label{sec:exp}
\subsection{Synthetic Data}
To test our algorithm, we created density functions as a convex combinations of densities from a chosen dictionary. The aim then was to reconstruct the density function from samples of the underlying random variable. We drew sample data from the synthetic density for various sample sizes ($N_s$) and tested the accuracy of the algorithms by averaging the absolute of log likelihood ratio between the estimated ($\widehat{f}_{\boldsymbol{X}}(.)$) and the known true PDF ($f_{\boldsymbol{X}}(.)$). We generated $M\triangleq 1000$ test samples $\{\boldsymbol{z}_k\}_{k=1}^M$ and used the following measure:
\begin{gather}\label{Likelihood}
    D(\widehat{f}_{\boldsymbol{X}}, f_{\boldsymbol{X}})\approx \frac{1}{M}\sum_{k=1}^{M}|\log(\widehat{f}_{\boldsymbol{X}}(\boldsymbol{z_k})/f_{\boldsymbol{X}}(\boldsymbol{z_k}))|.
\end{gather}
Notice that the marginalized distribution for the $n^{\textrm{th}}$ component of the RV will be $f_{X_n}(x_n)=\boldsymbol{\mathcal{D}_n}\boldsymbol{B_n}\boldsymbol{\lambda}$. This implies that by looking at the structure of the 1D empirical histogram (obtained from sample values) for each component, we can guess the families of distributions it might belong to. For example, the histogram of an exponential RV will peak at zero and decrease exponentially, that of a mixture of Laplacians will have distinct peaks at the mean of each component with a heavy tail, etc. 
For the parameters of the density families, we consider the range of values for samples of that component, say $[a,b]$. Consider that we wish to include Gaussians and Laplacians in our dictionaries. Then we can divide $[a,b]$ into regular intervals, and use the interval boundaries as the mean values for those distributions, while the variance/shape factors are chosen such that the densities with the mean values considered as described here, are sufficiently separated. 
In other words, we want the densities in our dictionary to completely cover the range of the samples for that component. If the $n^{\textrm{th}}$ component of the data is discrete, say label, categorical or integer data with $C_n$ states, we simply assign an identity matrix of size $C_n \times C_n$ to $\boldsymbol{\bar{D}_n}$. In this case the PMF of the marginal is given by: $f_{X_n}(x_n=i)=\sum_{r=1}^F \boldsymbol{\lambda}[r]\boldsymbol{B_n}[i,r]$. 

For all the experiments we compared our method with the PDF estimation algorithm from \cite{Kargas2019} and the well-known expectation maximization (EM) algorithm for GMM fitting \cite{Dempster1977}.  For EM-GMM, GMM-$n$ represents the performance plot for a GMM with $n$ clusters. In all experiments, we started with 5 clusters and went up to the point where increasing the number of clusters resulted in ill-conditioned co-variance. We refer to the algorithm in \cite{Kargas2019} as ``CPD-3". The technique described in \cite{Amiridi2020} was not included in the comparison results, as for many toy experiments it produced results similar to that of CPD-3 but was computationally very expensive for larger number of samples. We generated densities belonging to various families, as described below:

\noindent \textit{Mixture of Laplacians}: In the first experiment, we chose the dimension of the RV to be $N=5$ and $F=10$. Each column in the mode factor is a mixture of 5 Laplacian densities. Thus, if $\mathcal{L(\mu,\alpha)}$ denotes a Laplacian density with mean $\mu$ and shape factor $\alpha$, then each $f_{X_n|H}(x_n|H=r)=\sum_{i=1}^5w_{i,nr}\mathcal{L}(\mu_{i,nr},\alpha_{i,nr})$ where $\mu_{i,nr}\sim\mathcal{U}(-5,5)$ and $\alpha_{i,nr}\sim\mathcal{U}(1,2)$. Here and for all the following cases as well, the mixing weights $w_{i,nr}$ and $\lambda[r]\sim\mathcal{U}(0,1)$ and then $L_1$ normalized. Fig.~\ref{fig:exp1} shows $D(\widehat{f}_{\boldsymbol{X}}, f_{\boldsymbol{X}})$ vs number of samples $N_s$. Clearly, our method works better, be it in low or high sample regime, as Laplacian densities are neither smooth nor band-limited.

\noindent\textit{Mixture of Gaussians}: In the second experiment, we chose $N=6$ and $F=8$. Each column in the mode factor was chosen to be a mixture of 5 Gaussian densities. Thus, $f_{X_n|H}(x_n|H=r)=\sum_{i=1}^5w_{i,nr}\mathcal{N}(\mu_{i,nr},\sigma_{i,nr}^2)$ where $\mu_{i,nr}\sim\mathcal{U}(-5,5)$ and $\sigma_{i,nf}^2\sim\mathcal{U}(1,2)$. Fig.~\ref{fig:exp2} shows us that in lower sample regime, our algorithm is significantly accurate whereas in high sample regime it is not far from the best method. The reason for this performance anomaly is due to the fact that there is a limit to the accuracy with which the dictionary elements can represent a density due to parameter discretization while the algorithm CPD-3 performs sinc-interpolation which does not have any such restriction.

\noindent\textit{Mixture of Gaussians and Laplacians}: In this experiment, we chose $N=7$ and $F=10$. Each column in the first five mode factors is a mixture of 5 Laplacians while mixture of 5 Gaussian densities were used for next 5 mode factors. Again $\mu_{i,nr}\sim\mathcal{U}(-5,5)$ for both Gaussians and Laplacians, $\alpha_{i,nr}, \sigma_{i,nr}^2\sim\mathcal{U}(1,2)$. Just like the first case, our method performs significantly better than other algorithms across all sample regimes, as seen in Fig.~\ref{fig:exp3}. 

\noindent\textit{Mixture of Continuous and Discrete RVs}: For the last experiment, we chose $N=4$ and $F=8$. This time the last component in the RV was discrete with 10 states. As explained earlier, the dictionary $\boldsymbol{\bar{D}_4}$ was chosen to be a $10 \times 10$ identity matrix. Each column of the last mode factor $\boldsymbol{B_4}[:,r]$ was generated from $\mathcal{U}(0,1)$ and then $L_1$ normalized. Densities of the other mode factors were generated in the same way as for experiment 2. We could not compare our method with the EM-GMM algorithm because of its incapability to incorporate the discrete components of the RV in its formulation. The results for this experiment are presented in Fig.~\ref{fig:exp4}.

\subsection{Real Data}
\begin{figure}[t]
\centerline{\includegraphics[width=80mm,height=50mm]{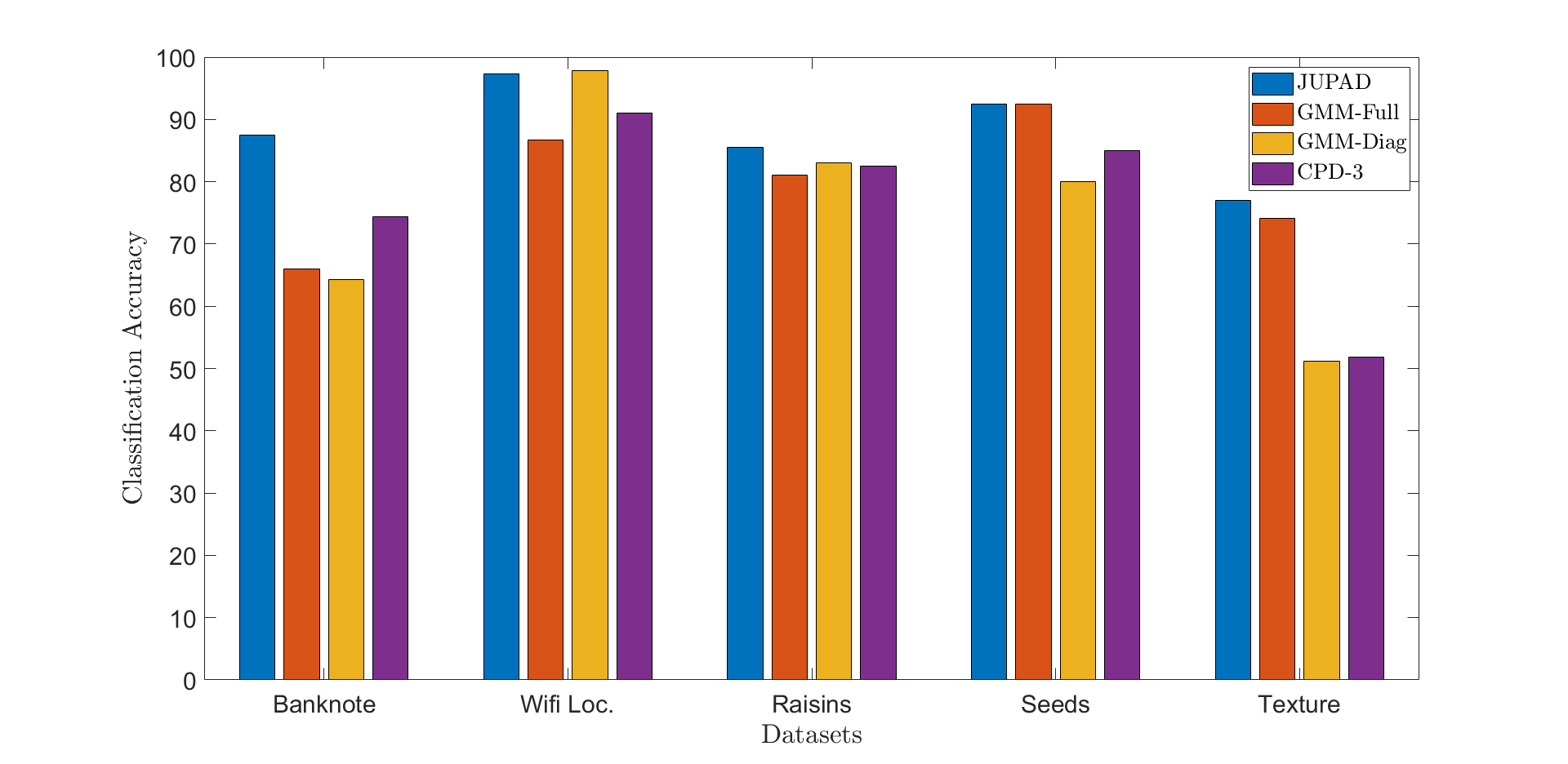}}
\caption{Classification accuracy on various datasets}
\label{fig:real_data}
\end{figure}
In the real-world datasets where the true underlying PDF is unknown, we tested our algorithm for a classification task on various datasets taken from UCI repository\footnote{\url{https://archive.ics.uci.edu/ml/datasets.php}} -- `Banknote Authentication' (5D), `Wifi Localization' (8D), `Raisins' (8D) and `Seeds' (8D) datasets, and the KTH TIPS texture dataset\footnote{\url{https://www.csc.kth.se/cvap/databases/kth-tips/download.html}}. The classification results for different methods are summarized in Fig.~\ref{fig:real_data}. The flexibility of our algorithm to adapt to hybrid distributions and learn the joint probability of both discrete and continuous components, endows us to estimate the joint density of the form $p_{\boldsymbol{X},Y}(\boldsymbol{x},y)$, where $\boldsymbol{X}$ denotes the vector of class feature (continuous) and $Y$ is their label (discrete). As described above, the dictionaries $\mathcal{D}_n$ for the features $\boldsymbol{X}$ are chosen from the continuous distribution families after examining their empirical marginals. For the label $Y$, $\mathcal{D}_n$ is set to identity matrix. The value of $F$ was chosen on the basis of accuracy on a validation set which was distinct from the training and test sets. For all the experiments, the classification task was performed by MAP estimation which assigns the label $\hat{y}=\textrm{argmax}_y p(y|\boldsymbol{x})=\textrm{argmax}_y p(\boldsymbol{x},y)/p(\boldsymbol{x})$. We ran the CPD-3 algorithm as implemented by the authors. Treating the label of each class as the latent variable, the classification in CPD-3 is done using the MLE estimate: $\textrm{argmax}_y p(\boldsymbol{x}|y)$, assuming the prior $p(y)$ to be uniform, unlike our algorithm where we learn the complete joint density $p(\boldsymbol{x},y)$. We also compared with GMM having full covariance (`GMM-Full') and diagonal covariance matrices (`GMM-Diag').

\noindent\textit{UCI Dataset}: Here we would like to bring out the generality of our method to model discontinuities in the PDFs. For eg., if the histogram of some component shows abrupt change at some value, then we can model this discontinuity by keeping few uniform distributions spanning the range of the histogram along with other densities. For most of the data, using a dictionary consisting of Gaussians and Uniform distributions yielded satisfactory results (see Fig.~\ref{fig:real_data}).

\noindent\textit{KTH TIPS}: This is a texture dataset which contains images of size $200\times200$ of various textures. We chose three textures - Orange Peel, Bread and Linen for our classification task. We used two training images from each class and divided them into $5\times5$ patches creating 26D (patch size+label) data. Thereafter, we normalized the pixel values so that all of them lie in the range $[0,1]$. Here, we used a dictionary consisting of only Gaussians. For testing, we created a collage of $5\times5$ patches of these textures and classified each patch again by using MAP estimator. Our model performed remarkably well for such a high-dimensional data and outperformed all the other algorithms by a reasonable margin (see Fig.~\ref{fig:real_data}).  
\section{Conclusion and Future Work}
We integrated ideas from low-rank tensors CPD and dictionary representation of signals to present a novel joint density estimation technique. Our method is completely general and can be applied to model mixtures of distributions coming from different families. The numerical results demonstrate the efficacy of our algorithm, especially in the low sample regime where other methods under-perform. Some future work may include a theoretical analysis of the proposed method w.r.t. sample complexity. 
Our method can also be extended to dictionary learning, where the dictionary elements are themselves learned from the data without any manual inspection.
\bibliographystyle{IEEEbib}
\bibliography{egbib}

\begin{thebibliography}{10}

\bibitem{Murphy2012}
Kevin~Patrick Murphy,
\newblock {\em Machine Learning: A Probabilistic Perspective},
\newblock MIT Press, 2012.

\bibitem{Wainwright2019}
G.~Young,
\newblock ``High‐dimensional statistics: A non‐asymptotic viewpoint, martin
  j. wainwright, cambridge university press, 2019, xvii 552 pages, £57.99,
  hardback isbn: 978‐1‐1084‐9802‐9,''
\newblock {\em International Statistical Review}, vol. 88, pp. 258--261, 04
  2020.

\bibitem{Chen2018}
Y.-C. Chen,
\newblock ``Lecture 6: Density estimation: Histogram and kernel density
  estimator,''
  \url{http://faculty.washington.edu/yenchic/18W\_425/Lec6\_hist\_KDE.pdf}.

\bibitem{Kargas2018}
N.~Kargas, N.~D. Sidiropoulos, and X.~Fu,
\newblock ``Tensors, learning, and ``kolmogorov extension'' for finite-alphabet
  random vectors,''
\newblock {\em IEEE Transactions on Signal Processing}, vol. 66, no. 18, 2018.

\bibitem{Ibrahim2021}
S.~Ibrahim and X.~Fu,
\newblock ``Recovering joint probability of discrete random variables from
  pairwise marginals,''
\newblock {\em IEEE Transactions on Signal Processing}, vol. 69, pp.
  4116--4131, 2021.

\bibitem{Vora2021}
J.~Vora, K.~S. Gurumoorthy, and A.~Rajwade,
\newblock ``Recovery of joint probability distribution from one-way marginals:
  Low rank tensors and random projections,''
\newblock in {\em IEEE Statistical Signal Processing}, 2021.

\bibitem{Kargas2019}
N.~Kargas and N.~D. Sidiropoulos,
\newblock ``Learning mixtures of smooth product distributions: Identifiability
  and algorithm,''
\newblock in {\em AISTATS}, 2019.

\bibitem{Amiridi2020}
M.~Amiridi, N.~Kargas, and N.~D. Sidiropoulos,
\newblock ``Low-rank characteristic tensor density estimation part {I}:
  Foundations,'' \url{https://arxiv.org/abs/2008.12315}.

\bibitem{NMF}
Daniel Lee and H.~Sebastian Seung,
\newblock ``Algorithms for non-negative matrix factorization,''
\newblock in {\em Advances in Neural Information Processing Systems}, T.~Leen,
  T.~Dietterich, and V.~Tresp, Eds. 2000, vol.~13, MIT Press.

\bibitem{Fu2018}
Xiao Fu, Kejun Huang, and Nicholas~D. Sidiropoulos,
\newblock ``On identifiability of nonnegative matrix factorization,''
\newblock {\em IEEE Signal Processing Letters}, vol. 25, no. 3, pp. 328--332,
  2018.

\bibitem{Gillis2014}
Nicolas Gillis and Stephen~A. Vavasis,
\newblock ``Fast and robust recursive algorithmsfor separable nonnegative
  matrix factorization,''
\newblock {\em IEEE Transactions on Pattern Analysis and Machine Intelligence},
  vol. 36, no. 4, pp. 698--714, 2014.

\bibitem{Kivinen1997}
Jyrki Kivinen and Manfred~K. Warmuth,
\newblock ``Exponentiated gradient versus gradient descent for linear
  predictors,''
\newblock {\em Information and Computation}, vol. 132, no. 1, pp. 1--63, 1997.

\bibitem{Beck2003}
Amir Beck and Marc Teboulle,
\newblock ``Mirror descent and nonlinear projected subgradient methods for
  convex optimization,''
\newblock {\em Operations Research Letters}, vol. 31, no. 3, pp. 167--175,
  2003.

\bibitem{Dempster1977}
A.~P. Dempster, N.~M. Laird, and D.~B. Rubin,
\newblock ``Maximum likelihood from incomplete data via the em algorithm,''
\newblock {\em JOURNAL OF THE ROYAL STATISTICAL SOCIETY, SERIES B}, vol. 39,
  no. 1, pp. 1--38, 1977.

\end{thebibliography}

\appendix
Here, we explain how we created the dictionary for PDF reconstruction using data inspection. We take three examples from the real-world dataset to show our method in detail. When we replace each mode factor by a linear combination of dictionary atoms, the marginalized density of $n^{\textrm{th}}$ component of the RV becomes:
\begin{gather}
    f_{X_n}(x_n)=\boldsymbol{\mathcal{D}_n}\boldsymbol{B_n}\boldsymbol{\lambda}
\end{gather}
where, $\boldsymbol{\mathcal{D}_n}$ is the continuous dictionary with various densities, $\boldsymbol{B_n}[:,r] \in \mathbb{R}_+^{L_n}\cup\{\boldsymbol{0}\}$ is the weight matrix with simplex constraint and $\boldsymbol{\lambda} \in \mathbb{R}_+^{F}$ is the weight vector for different mode factors again with simplex constriant.\\
The above form suggests that we can guess the columns of the continuous matrix $\boldsymbol{\mathcal{D}_n}$ by inspecting the histogram of $x_n$, the $n^{\textrm{th}}$ component of the RV. The method has already been described in the paper but for the sake of completeness we provide here some of the dictionaries we used for reconstructing joint PDF for real-world datasets.

\subsection{Seeds Dataset}
\begin{figure}[h]
\centerline{\includegraphics[width=70mm,height=40mm]{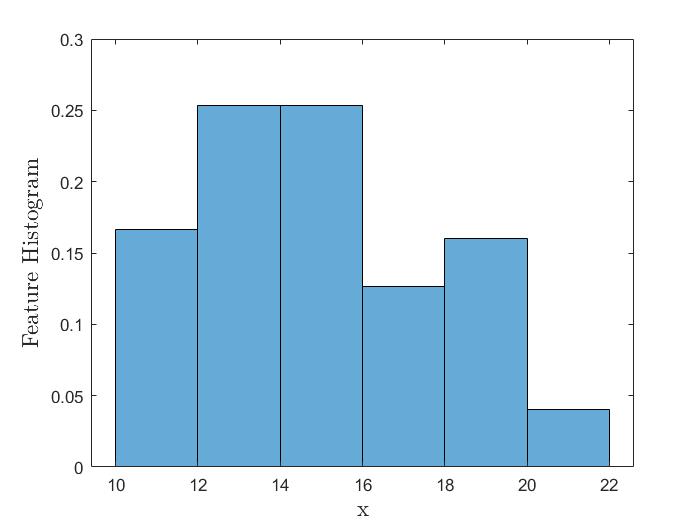}}
\caption{Histogram of a feature in seeds dataset}
\label{fig:seeds}
\end{figure}

Fig. \ref{fig:seeds} shows the histogram of a feature in seeds dataset. Clearly, such kind of histograms can never be accurately approximated by convex combination of only continuous densities. Thus to take into account the discontinuity at point 10, we included uniform pdfs in our dictionary. As explained in the paper, we divided the interval $[10,22]$ into small intervals of length 2 and the boundaries of these intervals were chosen as the mean for the Gaussians with variance 1, which ensured that these Gaussians are sufficiently apart across whole range. With this we included two uniform densities, $\mathcal{U}(10,16)$ and $\mathcal{U}(16,22)$.

\subsection{Wifi Localization Dataset}
\begin{figure}[h]
\centerline{\includegraphics[width=70mm,height=40mm]{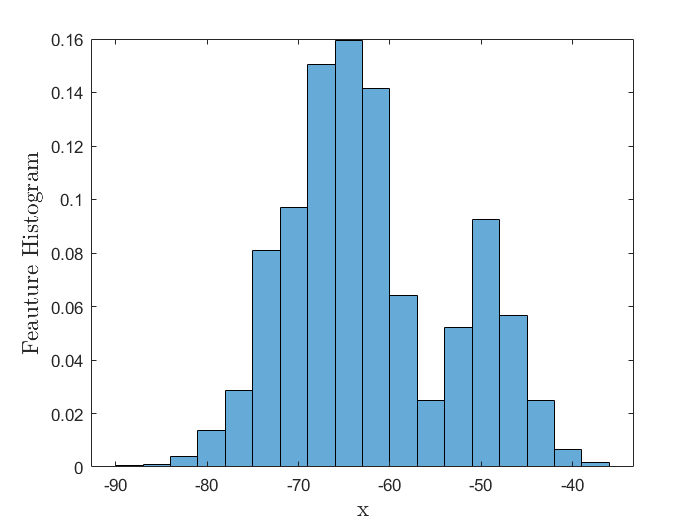}}
\caption{Histogram of a feature in wifi localization dataset}
\label{fig:wifi}
\end{figure}

Fig. \ref{fig:wifi} shows the histogram of a feature in wifi localization dataset. The histogram in this case clearly shows that the underlying pdf must have been convex combinations of Gaussians. Therefore, our dictionary atoms were Gaussians only. This time we divided the interval $[-90, -36]$ into intervals of length 4 and assigned boundary points as the means with variance equal to 4. 

\subsection{KTH TIPS Dataset}
\begin{figure}[h]
\centerline{\includegraphics[width=70mm,height=40mm]{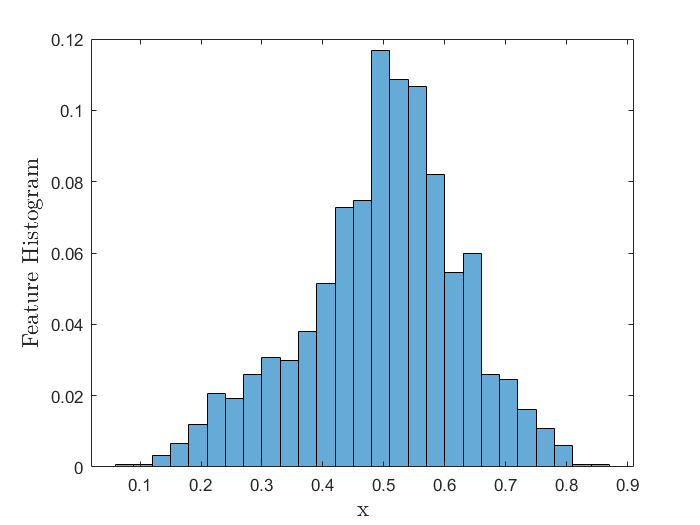}}
\caption{Histogram of a feature in texture dataset}
\label{fig:textures}
\end{figure}

Fig. \ref{fig:textures} shows the histogram of a feature in KTH TIPS dataset. Although it might look like for the 26D data (25 for patch size and 1 for label), inspecting each histogram and creating the dictionary is painstaking, all the 25 histograms (excluding label) had a similar structure and range because of feature normalization. Thus for all of them, the dictionary contained Gaussians with means separated at a distance of $0.04$ from each other and variance equal to $4\times10^{-4}$.
\end{document}